\pdfoutput=1

\documentclass[11pt]{article}

\usepackage[]{ACL2023}

\usepackage{times}
\usepackage{latexsym}
\usepackage{amsmath}
\usepackage{graphicx}
\usepackage{booktabs}
\usepackage{multirow}
\usepackage{color, colortbl}
\usepackage{cleveref}
\usepackage{subcaption}
\usepackage{pifont}
\newcommand{\cmark}{\ding{51}}%
\newcommand{\xmark}{\ding{55}}%

\definecolor{LightCyan}{rgb}{0.88,1,1}

\usepackage[T1]{fontenc}

\usepackage[utf8]{inputenc}

\usepackage{microtype}

\usepackage{inconsolata}
\usepackage{todonotes}

\title{Learning from Children: \\
Improving Image-Caption Pretraining via Curriculum}

\author{Hammad A. Ayyubi\textsuperscript{1}, Rahul Lokesh\textsuperscript{1}, Alireza Zareian\textsuperscript{2}, Bo Wu\textsuperscript{3}, Shih-Fu Chang\textsuperscript{1}
\\
\small \textsuperscript{1} Columbia University \quad
\small \textsuperscript{2} Snap Inc. \quad
\small \textsuperscript{3} MIT-IBM Watson AI Lab \quad \\
\small \texttt{\{ha2578, rl3164\}@columbia.edu}
}

\begin{document}
\maketitle
\begin{abstract}

Image-caption pretraining has been quite successfully used for downstream vision tasks like zero-shot image classification and object detection. 
However, image-caption pretraining is still a hard problem -- it requires multiple concepts (nouns) from captions to be aligned to several objects in images. 
To tackle this problem, we go to the roots -- the best learner, children. We take inspiration from cognitive science studies dealing with children's language learning
to propose a curriculum learning framework.
The learning begins with easy-to-align image caption pairs containing one concept per caption. 
The difficulty is progressively increased with each new phase by adding one more concept per caption. 
Correspondingly, the knowledge acquired in each learning phase is utilized in subsequent phases to effectively constrain the learning problem to aligning one new concept-object pair in each phase.
We show that this learning strategy improves over vanilla image-caption training in various settings -- pretraining from scratch, using a pretrained image or/and pretrained text encoder, low data regime etc. Code available at: \url{https://github.com/hayyubi/cur_vl.git}.

\end{abstract}

\begin{figure}[t]
    \centering
    \captionsetup{skip=5pt}
    \includegraphics[width=\linewidth]{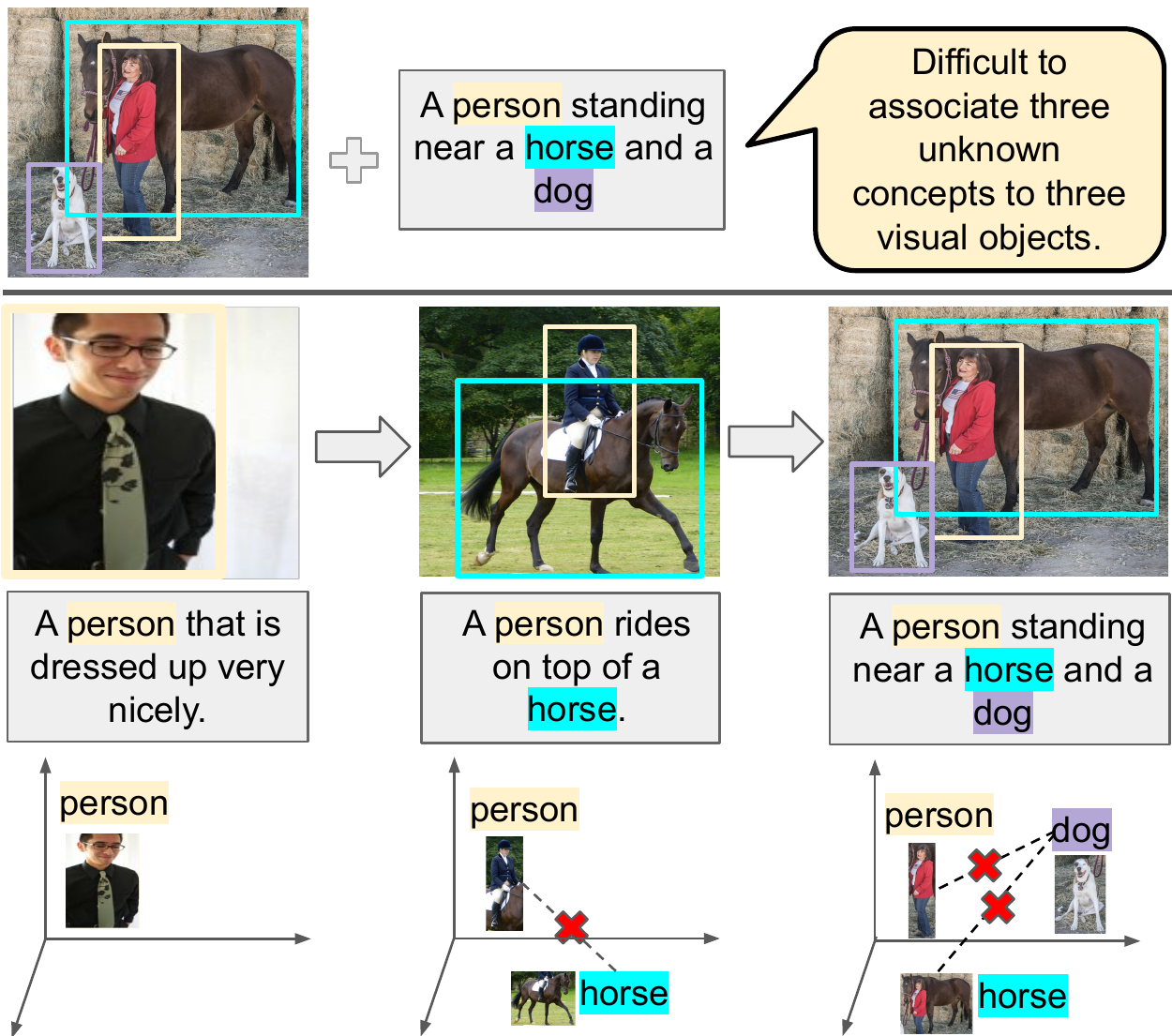}
    \caption{Top: Normal image-caption pertaining. Bottom: Proposed Curriculum Learning Framework. The curriculum eases the learning problem by requiring the model to align only one concept-object pair at a time.}
    \label{fig:cur_main}
\end{figure}

\section{Introduction}

Recently, there has been a tremendous interest in 
employing image-caption pretraining for downstream vision tasks like zero-shot object classification \cite{clip} and zero-shot object detection \cite{zareian-2021-ovr, glip}. The idea is to learn a common semantic space where the visual embeddings of objects in images lie close to the textual embeddings of the concepts (objects' name/tag/label) in captions they refer to. This learned semantic space is later exploited 
for zero-shot object recognition
by finding the concept embedding nearest to the objects' embeddings.

Despite the recent success, image-caption pretraining is a complex problem as it entails aligning multiple concepts in a caption with multiple objects in an image, as shown in \cref{fig:cur_main}. Different methods have tried to solve this problem from various angles -- CLIP \cite{clip} and ALIGN \cite{align} by using more data, ALBEF \cite{albef} by using more complex network architecture, Florence \cite{florence} and CoCa \cite{coca} by using more tasks and ERNIE-ViL 2.0 \cite{ernie-vil-2.0} by using more data augmentations (views).

We propose an alternative approach based on a novel learning strategy that is architecture agnostic and does 
not require any additional data or compute. 
We take inspiration from cognitive science research studying how children learn language (concepts) in early stages by just observing their surroundings (images). 
Specifically, we refer to two studies
showing that childern learn rapidly if the object of interest is unambiguous \cite{toddler2014} and by applying co-referential statistics across multiple scenes \cite{Smith2008InfantsRL}.

We implement these two ideas via a curriculum learning approach (demonstrated in \cref{fig:cur_main}):
\begin{enumerate}
    \item We train the model in multiple phases of increasing difficulty with each phase containing one more concept in the caption than the previous one. Moreover, each phase contains only one new concept, the rest seen in prior phases.
    \item In each phase, we leverage the concept-object association learned in prior phases to recognize the seen concepts and focus on aligning the new/unseen concept (\cref{subsubsec: caal}).
\end{enumerate}
These two strategies effectively reduce the problem of aligning multiple object-concept pairs per training sample to aligning only one such pair.

To the best of our knowledge, no prior work has applied curriculum leaarning to image-caption pretraining in this way. \citet{curriculum-neg-samples} apply a curriculum based on the difficulty of negative samples in contrastive loss. Whereas, \citet{liu-etal-2021-msp} design the curriculum based on the granularity of text: from words to phrases to sentences.

Although our proposed approach can be applied to any multimodal network architecture, we pick OVR-CNN \cite{zareian-2021-ovr} due to its simplicity. 
We pretrain it with the proposed curriculum learning approach and evaluate on the downstream task of zero-shot object detection. 
We demonstrate that curriculum learning outperforms vanilla image-caption pretraining on a variety of architectural settings -- with and without a pretrained image encoder and/or a pretrained text encoder. 
We even show superior performance in low-data settings, suggesting our method can be leveraged in low-resource scenarios as well.

\begin{figure*}
\captionsetup{skip=3pt}
\begin{subfigure}[t]{0.32\linewidth}
\captionsetup{skip=0pt}
\includegraphics[width=\linewidth]{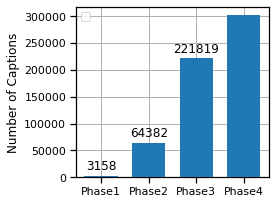}
\caption{}
\label{fig:cur_stat_numcapVphase}
\end{subfigure}
\hfill
\begin{subfigure}[t]{0.33\linewidth}
\captionsetup{skip=0pt}
\includegraphics[width=\linewidth]{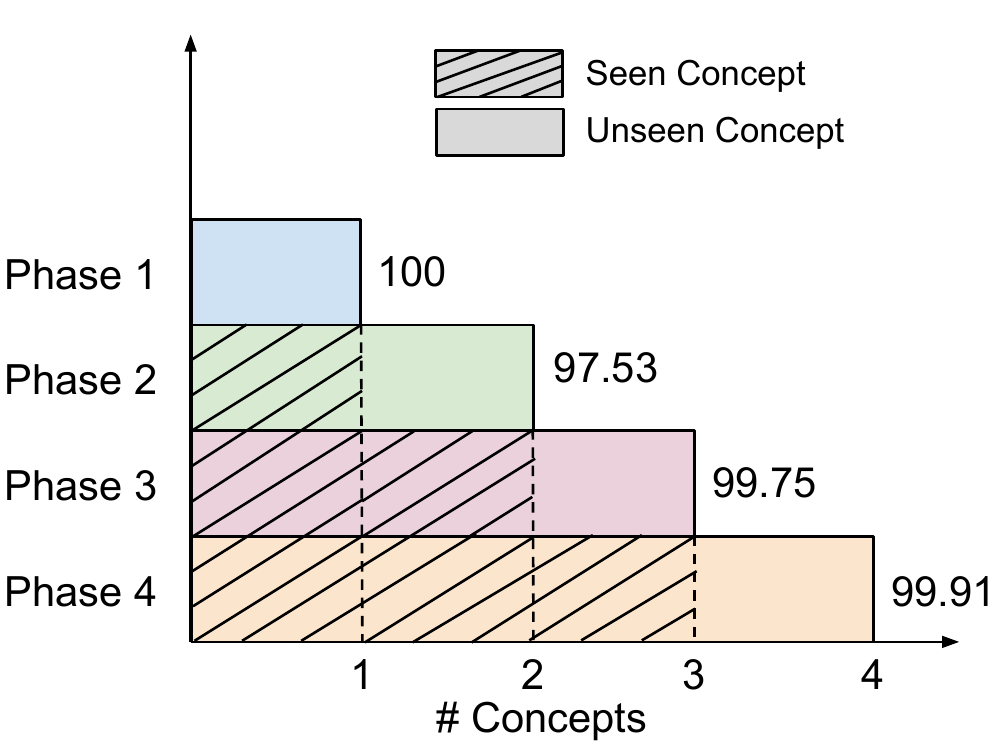}
\caption{}
\label{fig:cur_stat_prevNoun}
\end{subfigure}
\hfill
\begin{subfigure}[t]{0.33\linewidth}
\captionsetup{skip=0pt}
\includegraphics[width=\linewidth]{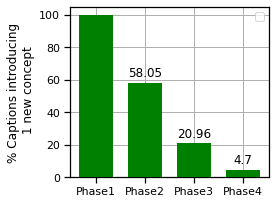}
\caption{}
\label{fig:cur_stat_intro1concept}
\end{subfigure}
\caption{Curriculum Statistics. (a) \#Captions Vs Phase (b) The number next to bar shows \% of captions per phase with at least \#shaded concepts seen previously. (c) \% Captions per phase introducing 1 new concept}
\end{figure*}


\section{Method}
We propose a curriculum learning framework to improve image caption pretraining. In this work, we apply it to OVR-CNN as its architecture is simpler and easier/faster to train/evaluate. We begin the description of our approach with a brief background on OVR-CNN. Next, we discuss how we modify it to implement the proposed curriculum learning framework.

\subsection{OVR-RCNN Background}
OVR-RCNN is a dual-encoder (separate visual encoder and text encoder) multimodal architecture. First, it pretrains the encoders using image-captions and later utilizes them for the downstream task of object detection. We only discuss the pretraining procedure as we only utilize this component.

OVR-RCNN's visual encoder is ResNet-50 \cite{resnet} and text encoder is either BERT \cite{devlin-etal-2019-bert} or GloVE \cite{pennington-etal-2014-glove}. The visual encoder takes an image, $I$ with $w\times h$ dimensions, as input and outputs a feature map of $w/32\times h/32$ regions. Each feature map is a vector which is transferred to language space using a projection layer. This gives the visual embeddings, $e^I_i$, for each region $i$. The tokenized caption, $C$, is input to the text encoder which outputs an embedding $e^C_j$ for each token $j$.

The token-image region pair is aligned via weak supervision. Specifically, a global alignment score between image and caption, $\langle I,C \rangle_G$ is calculated using a locally weighted average alignment score of image regions and tokens as follows:

\begin{equation}
\label{eq:align_score}
\small
    \langle I,C \rangle_G = \frac{1}{n_C} \sum_{j=1}^{n_C} \sum_{i=1}^{n_I} a_{i,j} \langle e_i^I, e_j^C \rangle_L
\end{equation}
where $\langle ., . \rangle_L$ is the dot product of two vectors, $n_I$ and $n_C$ are the number of image and caption tokens respectively, and
\begin{equation}
\label{eq:attn}
\small
a_{i,j} = \frac{\exp \langle e_i^I, e_j^C \rangle_L}{\sum_{i'=1}^{n_I} \exp \langle e_{i'}^I, e_j^C \rangle_L} 
\end{equation}
The model is trained using contrastive learning by maximizing the global alignment score, $\langle I,C \rangle_G$, between positive image-caption pairs and minimizing it between negative pairs sampled from the same training batch.
\begin{equation}
\label{eq:loss}
\resizebox{0.89\hsize}{!}{$
    \mathcal{L} = - \log \frac{\exp \langle I, C \rangle_G}{\sum_{\{I',C'\} \in \mathcal{N}_{I,C}} \exp \langle I', C' \rangle_G + \exp \langle I, C \rangle_G}
    $}
\end{equation}
where, $\mathcal{N}_{I,C} = \{I,C'|C'\in\mathcal{B}_C\} \cup \{I',C|I'\in\mathcal{B}_I\} $ and $\mathcal{B}_C, \mathcal{B}_I$ are batch captions and batch images respectively. This learning objective aligns paired image and caption together and also provides weak supervision for image-regions and caption-tokens association. 

\subsection{Curriculum Learning Framework}
\label{subsec:cur}
OVR-CNN facilitates object-concept alignment through coarse image-region and concept alignment.
However, as an object can span multiple image regions or multiple objects can span an image region, this strategy can be noisy. 
To eliminate this noise and focus on the contribution of our curriculum framework to object-concept alignment, we train the model using object region features instead of image region features.
To this end, object region bounding box is used to ROI pool \cite{girshickICCV15fastrcnn} the image region features. The resulting feature vector $e^I_o$, for each object $o$, is used to replace $e^I_i$ in \cref{eq:align_score,eq:attn}.


\subsubsection{Curriculum Design}
\label{subsubsec: curriculum}

The learning is divided into $1,2,3\dots k$ phases. Each phase $p$ is trained with only those image-caption pairs having $p$ concepts per caption. To divide the data into phases, we use spacy\footnote{\url{https://spacy.io/usage}} to PoS (Part of Speech) tag the captions. Depending upon the number of nouns in each caption, the caption and its paired image is grouped into the corresponding phase. 
This strategy of designing the curriculum also imparts the data an additional property empirically -- at most only one new concept is introduced per caption in each phase (as demonstrated in \cref{fig:cur_stat_prevNoun}).

\subsubsection{Curriculum Aware Alignment Loss}
\label{subsubsec: caal}
To recognize the concepts in captions previously seen in prior phases and focus on aligning the new/unseen concept, we formulate a novel Curriculum Aware Alignment Loss ($\mathcal{L_C}$). Specifically, we first calculate the previously learned object-concept alignment, $a_{o,j}$ from modified \cref{eq:attn}, using either the trained model from the last iteration ($\mathcal{L_{CR}}$) or the trained model from the last phase ($\mathcal{L_{CP}}$). Next, $a_{o,j}$ is used to compute:
\begin{equation*}
    \resizebox{1\hsize}{!}{$
    a_{o,j}^{'} = \frac{\exp \langle e_o^I, e_j^C \rangle_L \exp{(-\max_o(a_{o,j}).\frac{t}{T})}}{\sum_{o'=1}^{n_I} \exp \langle e_{o'}^I, e_j^C \rangle_L \exp{(-\max_o(a_{o,j}).\frac{t}{T})}} 
    $}
\end{equation*}
where, $t$ is the current iteration number and $T$ is the total number of iterations in training. 

For a concept $j$, which is already closely aligned to an object $o$, $\max_o(a_{o,j})$ is high. This leads to a low value of $a_{o,j}^{'}$, resulting in less attention being paid to concept $j$ in the current training iteration/phase. Vice versa for a concept that is not well aligned with any object. $a_{o,j}^{'}$ effectively redistributes the attention of learning to focus more on concepts that are not well aligned with any object. The term $t/T$ has a low value in the beginning of training and gradually scales to $1$ by the end. This allows the network to ignore prior knowledge in the beginning while utilizing it in the latter stages.

We use $a_{o,j}^{'}$ to replace $a_{o,j}$ in modified \cref{eq:align_score}, and then use \cref{eq:loss} to compute $\mathcal{L_C}$.

\section{Experiments}


\subsection{Pretraining Dataset and Implementation Details}
We use the COCO Captions dataset \cite{chen2015microsoft} for pretraining. It contains 118,287 images and 5x captions.
To obtain bounding box regions for objects in images, we use COCO Objects \cite{lin2014microsoft} dataset as it uses the same set of images as COCO Captions.
We divide the data into $k=4$ phases using the strategy discussed in \cref{subsubsec: curriculum}. \Cref{fig:cur_stat_numcapVphase} shows number of captions assigned to each phase. 
As shown in fig. 2b, the majority of captions in each phase have at least k-1 concepts previously seen, allowing the curriculum to introduce at most one new concept per training sample.
Further, as more concepts are introduced with each passing phase, the percent of captions per phase actually introducing a new concept decreases (as depicted in \cref{fig:cur_stat_intro1concept}).
By phase 4, this percent reduces to $<5\%$. Additional phases of training may not contain enough captions actually introducing a new concept in the curriculum way, making these phases similar to regular image-caption training. Hence, we limit to 4 phases.

We train the model using SGD optimizer, with a batch size of 32 for 4 epochs in each phase, a learning rate of 0.005, a step scheduler, and the loss $\mathcal{L_{CP}}$.

\subsection{Downstream Task, Dataset and Transfer}
\label{subsec:downstream}
We evaluate the performance of the model on zero-shot object detection task on COCO Objects, val split (4836 images; 33374 instances of 65 object class). The task involves object bounding box predictions besides classifying these object regions to a label (concept). However, our method is aimed only at improving the alignment of object regions to a concept. As such, we eliminate any performance noise from bounding box predictions by only evaluating the classification accuracy of object regions given ground truth object bounding boxes. 

Transfer to Downstream Task: 
We extract object features from image and object bounding box using visual backbone and use it to find the closest class label vector (obtained via language backbone).

\begin{table}[t]
\centering
\captionsetup{skip=3pt}
\fontsize{9}{11}\selectfont
\begin{tabular}{lccc}
\toprule
\multirow{2}*{Model}      & Pretrained  & Language   & \multirow{2}*{Accuracy} \\
& Visual BB & Backbone & \\
\midrule
OVR-CNN$_O$ & \xmark & GloVE  & 20.62$_{\pm 0.86}$  \\
\rowcolor{LightCyan}
Ours & \xmark & GloVE  & \textbf{21.64$_{\pm 1.02}$}\\
\midrule
OVR-CNN$_O$ & \xmark & BERT & 22.73$_{\pm 0.06}$  \\
\rowcolor{LightCyan}
Ours & \xmark & BERT  & \textbf{23.74$_{\pm 0.48}$}\\
\midrule
OVR-CNN$_O$ & \cmark & BERT & 34.46$_{\pm 0.11}$  \\
\rowcolor{LightCyan}
Ours & \cmark & BERT  & \textbf{35.49$_{\pm 0.21}$}\\
\bottomrule
\end{tabular}
\caption{Curriculum learning vs. baseline in various settings with/without pretrained encoders. BB: backbone}
\label{tab:main-result}
\end{table}

\begin{figure}
\centering
  \begin{minipage}[t]{0.55\linewidth}
    
\centering
\captionsetup{skip=3pt}
\fontsize{8}{11}\selectfont
\setlength{\tabcolsep}{1.5pt}
\begin{tabular}{lcc}
\toprule
Model     & $P_1$ Obj. & $P_2$ Obj. \\
\midrule
OVR-CNN$_O$  & 49.36 & 16.61  \\
Ours & \textbf{50.9} & \textbf{26.1}\\
\bottomrule
\end{tabular}
\captionof{table}{Phase wise top-5 accuracy. $P_i$ Obj.: Phase $i$ Objects.}
\label{tab:phase-wise-acc}

  \end{minipage}
  \hfill
  \begin{minipage}[t]{0.405\linewidth}
    
\centering
\captionsetup{skip=3pt}
\fontsize{8}{11}\selectfont
\setlength{\tabcolsep}{0.1pt}
\begin{tabular}{lc}
\toprule
Model     & Accuracy \\
\midrule
OVR-CNN$_O$  & 13.10  \\
Ours & \textbf{17.45}\\
\bottomrule
\end{tabular}
\captionof{table}{Object bounding boxes from RPN.}
\label{tab:rpn_tabular}

  \end{minipage}
\end{figure}

\begin{figure}[t]
    \centering
    \captionsetup{skip=-1.5pt}
    \includegraphics[width=0.9\linewidth]{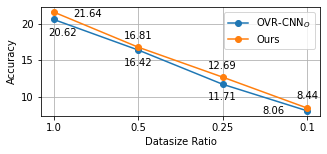}
    \caption{Performance comparison in low data setting.}
    \label{fig:low-data}
    \vspace{-1,8pt}
\end{figure}
\begin{table}[t]
\centering
\captionsetup{skip=3pt}
\fontsize{9}{11}\selectfont
\begin{tabular}{lccc}
\toprule
Model      & Cur.  & Loss   & Accuracy \\
\midrule
OVR-CNN$_O$ &  & $\mathcal{L}$  & 19.73  \\
OVR-CNN$_O$ &  & $\mathcal{L_{CR}}$  & 20.65  \\
\midrule
Ours & \cmark & $\mathcal{L}$  & 21.02\\
Ours & \cmark & $\mathcal{L_{CR}}$  & 21.58\\
Ours & \cmark & $\mathcal{L_{CP}}$  & \textbf{22.57}\\
\bottomrule
\end{tabular}
\caption{Ablation of proposed loss.}
\label{tab:loss-ab}
\end{table}

\subsection{Baseline and Evaluation}
Our baseline is OVR-CNN, a regular image-caption pretrained model. 
However, since our method uses object region features instead of image patch features for multimodal alignment (\cref{subsec:cur}), we also pretrain OVR-CNN with object regions to obtain OVR-CNN$_O$. It is transferred to downstream task similar to our proposed model (\cref{subsec:downstream}).

Our proposed curriculum framework outperforms the baseline in various settings, as shown in \cref{tab:main-result}. The accuracy numbers reported are averaged across three seeds. This demonstrates that our proposed learning strategy works across encoders trained from scratch or pretrained ones.

\textbf{Performance Gain Analysis.} 
We analyze model performance on object classes introduced during pretraining in phase 1 and phase 2 separately.
As reported in \cref{tab:phase-wise-acc}, the improvement in phase 2 objects is \textasciitilde 10x. 
This illustrates that our curriculum strategy improves alignment of multiple concepts in a caption by focusing on one at a time.

\textbf{Low Data Setting.} 
Our model outperforms the baseline even if both uses 50\%, 25\% or 10\% data (\cref{fig:low-data}), indicating its utility when data is scarce.

\textbf{Region proposals instead of ground-truth object regions.} We use a RPN model \cite{fast-rcnn} trained class-agnostically on Visual Genome \cite{krishnavisualgenome} to generate object regions. The superior performance of our model against baseline, reported in \cref{tab:rpn_tabular}, demonstrates that 
our approach is effective even when ground-truth object regions are not available.

\textbf{Loss Ablation.} From \cref{tab:loss-ab}, we can conclude that our curriculum design works (Ours + $\mathcal{L}$ > OVR-CNN$_O$ + $\mathcal{L}$); our proposed curriculum aware loss works (Ours + $\mathcal{L}$ < Ours + $\mathcal{L_{CR}}$) irrespective of curriculum (OVR-CNN$_O$ + $\mathcal{L}$ < OVR-CNN$_O$ + $\mathcal{L_{CR}}$); curriculum aware loss works better when previous knowledge is taken from the last phase instead of the last iteration
(Ours + $\mathcal{L_{CP}}$ > Ours + $\mathcal{L_{CR}}$).

\begin{figure*}[t]
    \centering
    \captionsetup{skip=7pt}
    \includegraphics[width=\linewidth]{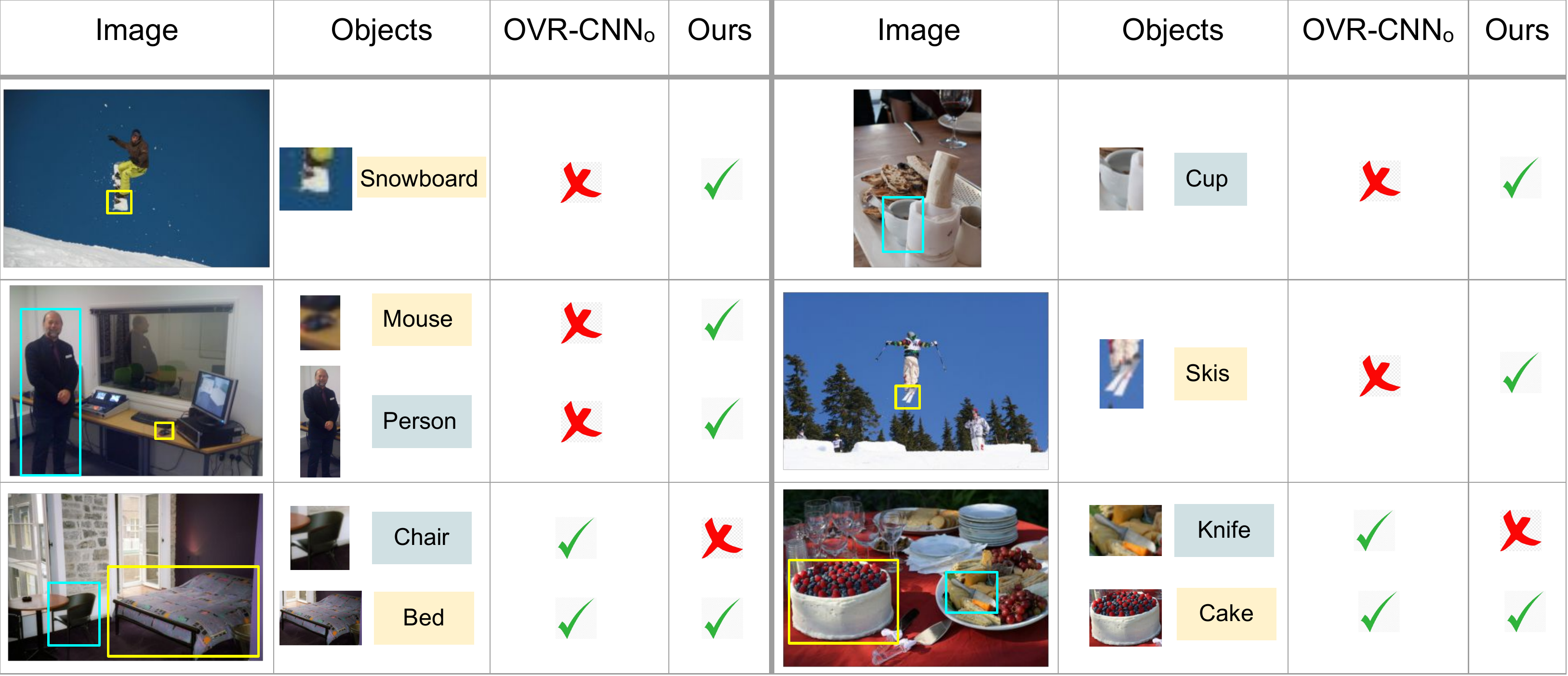}
    \caption{Qualitative analysis of cases where our approach works/doesn't work. The top two rows show samples where our model is successfully able to align the objects with the correct concept while OVR-CNN$_O$ makes mistakes. Interestingly, most of these objects are from Phase 2 -- "snowboard", "cup", "skis" etc. The bottom row shows cases where our model makes mistakes. Note: only relevant, not all, objects are shown from each image. }
    \label{fig:qual_analysis}
\end{figure*}

\textbf{Qualitative Analyssis.} We provide qualitative analysis as well to shed more insights into the cases where our approach works/doesn't work. From \Cref{fig:qual_analysis}, we find that our model performs better than OVR-CNN$_O$ in certain cases, especially when the objects are from Phase 2 -- "snowboard", "cup", "skis" etc. This provides further evidence towards our claim that our approach improves the alignment of Phase 2 objects.

\textbf{Comparison of traditional mAP metric for object detection}
\begin{table}[t]
\centering
\captionsetup{skip=5pt}
\fontsize{9}{11}\selectfont
\begin{tabular}{lccccc}
\toprule
\multirow{2}*{Model}      & \multirow{2}*{Base}  & \multirow{2}*{Target}   & \multicolumn{3}{c}{Generalized} \\
&&& Base & Target & All \\
\midrule
OVR-CNN  & 46.8 & \textbf{27.5} & 46.0 & \textbf{22.8} & 39.9  \\
Ours & \textbf{48.89} & 26.65 &\textbf{48.18} & 21.99 & \textbf{41.33}\\
\bottomrule
\end{tabular}
\caption{Comparison of traditional mAP metric for object detection.}
\label{tab:map}
\end{table}
As mentioned before, we have focused our experiments on evaluating object-concept alignment rather than on traditional object detection mAP metric. This was done to avoid unnecessary performance noise arising from training a RPN, which is required for mAP evaluation. However, to test the limits of our model, we evaluate on this noisy mAP metric as well. We keep all the settings similar as \citet{zareian-2021-ovr}, except we pretrain using our curriculum learning approach. The results are reported in \Cref{tab:map}. We find that our model performs better in the most generic Generalized (`All') set (41.33 vs 39.9), signifying the effectiveness of our approach even in this noisy setting. We further observe that we perform better in the base classes while lagging behind in the target classes. A deeper analysis shows that most of the Phase 2 objects, on which we make major improvements, lie in the base classes. This explains the improved performance on base classes and slight depreciation in target classes performance.

\textbf{Training with image grid regions.}
\begin{table}[t]
\centering
\captionsetup{skip=5pt}
\fontsize{9}{11}\selectfont
\begin{tabular}{lc}
\toprule
Model     & Accuracy \\
\midrule
OVR-CNN  & \textbf{23.45}  \\
Ours & 23.33\\
\bottomrule
\end{tabular}
\captionof{table}{Training with image grid regions instead of object regions.}
\label{tab:region}
\end{table}
Our curriculum based pretraining method was aimed at improving object-concept alignment by focussing on one object at time. To facilitate this, we pretrained directly with object regions. Image regions were not used to eliminate noise arising from an object spanning multiple regions or multiple objects being present in the same image region (object presence noise). However, we further push the limits of our model to assess how it performs when trained with noisy image regions instead of object regions. The results are reported in \Cref{tab:region}. We find that our model performs slightly worse than OVR-CNN (23.33 vs 23.45). We attribute this performance degradation to the inherent object presence noise in image regions as discussed earlier. 

\vspace{-5pt}
\section{Conclusion}
We proposed a curriculum learning framework to improve image-caption pretraining, using the number of concepts in captions. We also designed a novel curriculum aware loss to focus learning on the unaligned concept in each phase. Our approach outperforms vanilla image-caption pretraining in various settings, including with/without pretrained encoders and small data. Further, we extensively analysed our model to study the contribution of each component.

\section*{Limitations}
Although our proposed curriculum can be applied to any multimodal architecture, curriculum aware loss requires modifications for use with dual encoder architectures that don't use cross-modal attention. Additionally, we use an off-the-shelf Part-of-Speech tagger to divide the data into different phases. As such, the correctness of this division is dependent on the quality of tagger. A poor tagger can negatively impact the curriculum design. Moreover, our approach doesn't apply to possible image-captions dataset which contain only short captions, containing possibly only one noun.

\section*{Acknowledgement}
This work was supported by the U.S. DARPA GAILA Program No.HR00111990058. The views
and conclusions contained in this document are those of the authors and should not be interpreted as
representing the official policies, either expressed or implied, of the U.S. Government. The U.S. Government is authorized to reproduce and distribute reprints for Government purposes notwithstanding
any copyright notation here on.


\bibliography{anthology,custom}
\bibliographystyle{acl_natbib}




\end{document}